\documentclass[pdflatex,sn-mathphys-num]{sn-jnl}

\usepackage{graphicx}%
\usepackage{multirow}%
\usepackage{amsmath,amssymb,amsfonts}%
\usepackage{amsthm}%
\usepackage{mathrsfs}%
\usepackage[title]{appendix}%
\usepackage{xcolor}%
\usepackage{textcomp}%
\usepackage{manyfoot}%
\usepackage{booktabs}%
\usepackage{algorithm}%
\usepackage{algorithmicx}%
\usepackage{algpseudocode}%
\usepackage{listings}%
\usepackage{url}%
\usepackage{placeins}%
\usepackage{float}%

\theoremstyle{thmstyleone}%

\theoremstyle{thmstyletwo}%

\theoremstyle{thmstylethree}%

\usepackage{url}
\begin{document}

\title[Verifiable Blind Evaluation of LLM Judges]{Who Verifies the Benchmark? Decentralizing Trust in Large Language Model Evaluation}


\author[1]{\fnm{Sahil} \sur{Pardasani}}\email{sahil.pardasani@gmail.com}

\author*[1]{\fnm{Madhusudan} \sur{Singh}}\email{mps6990@psu.edu}

\affil*[1]{\orgdiv{Blockchain Data Intelligence Lab }
\orgname{The Pennsylvania State University}, \orgaddress{\street{201 Old Main}, \city{University Park}, \postcode{16802}, \state{PA}, \country{United States of America}}}


\abstract{Large Language Model benchmarking showcases organizations' expertise to attract customers. Unverified DeepSeek R1 benchmark claims against OpenAI's o-1 model caused a market panic and $589$ billion in NVIDIA losses on January 27, 2025. Without encryption, vendor-led honor systems support benchmark claims. Academic re-evaluations and third-party leaderboards show systemic benchmark manipulation, including secret proprietary model alterations, contaminated training corpora, and selective disclosure, beyond market volatility. LLM-as-a-judge replaces human assessment for scalability. According to studies, LLMs may have identity-aware bias, which adjusts evaluation rates depending on response model rather than answer quality. No study has measured or rectified identity prejudice in politically sensitive, reasoning-intensive, and preference-based information. Seven model verifiers (GPT OSS 120B, Llama 3.3 70B, GLM 5.1, Qwen3 32B, DeepSeek V4 Pro, Mistral Large 3, Sarvam M) anonymously and transparently score three primary models' responses on a 58-question benchmark for factual, reasoning, political, and preference-based categories to examine identity-aware bias Identity disclosure increases scores somewhat for factual questions, moderately for stress-reasoning tasks, and significantly for geopolitically sensitive topics (GLM 5.1: +7.00 points, p = 0.0249; Llama 3.3 70B: +1.56 points, p = 0.00 We provide a blockchain-based commit-reveal protocol employing Autonomous Economic Agents (AEAs) on an Ethereum-compatible ledger to confine judge scores before disclosing candidate names based on empirical data. Phase 1 commits a one-way hash of the score and secret salt to the blockchain before disclosing identity, and Phase 2 exposes identity and submits raw score for on-chain verification. This provides an auditable evidence trail that separates blind evaluation from post-hoc statements, decreasing independent researcher and third-party leaderboard verification responsibilities.}

\keywords{Artificial Intelligence, Benchmarking, Blockchain, Large language model (LLM)}

\maketitle

\section{Introduction}\label{sec1}

LLM benchmarks have become like a primary currency of AI progress and a systematically gamed measurement system whose integrity rests on uncompensated academic labor. Self-reported scores from various big companies move billion-dollar markets. The most notable such instance being on January 27, 2025, when DeepSeek R1 benchmark claims against OpenAI's o-1 model triggered a panic that erased \$589 billion from NVIDIA's market capitalization in a single day. This was fueled by the claim that DeepSeek matched OpenAI's model's reasoning capabilities, for instance scoring 79.8\% on the AIME 2024 math benchmark as compared to 79.2\% (by ChatGPT o-1) while spending less than \$6 million on training using hardware restricted by US export controls~\cite{bib14}.

 One of the largest single-day losses in the USA stock market, primarily driven by tech stocks was triggered by this as investors panicked. Even after such high stakes movement in stock prices which led to no framework exists to make sure these benchmarking claims ar right before their release. A major incident that appeared in mainstream media was that of Meta's Llama 4 LMArena submission which was then even confirmed by ex Meta Chief Scientist Yann LeCunn in an interview with Financial Times~\cite{bib19}. Yet another such instance was that of OpenAI funding the leading benchmark called FrontierMath~\cite{bib22}.Our work uses this as a backdrop to propose a blockchain-based verification infrastructure capable of anchoring these claims in cryptographic truth rather than vendor-led honor systems. We propose a decentralised verification protocol which utilizes Autonomous Economic Agents (AEAs) and a blockchain-based commit-reveal mechanism. By cryptographically binding judge commitments before identity disclosure, our protocol replaces voluntary claims with mathematical enforcement for LLM evaluation.

The sections below discuss about the following: related work/literature review, methodology, experimentation, mathematics, system architecture diagram, experimental analysis, proposed solution, discussion, conclusion, and references.

\section{Related Work / Literature Review}\label{sec2}

There have been recent findings that biases exist across several families of models, and they shift increasingly towards the negative when the candidate models' origins or identities are known~\cite{bib1,bib2}.

One striking observation is that when we know that the judge and candidate belong to the same model family, scores tend to be inflated, termed self-bias. A 2025 study found that models like ChatGPT-4o and Claude 3.5 Sonnet assigned higher scores to their own outputs and also displayed family bias~\cite{bib2}.

Another such scenario is seen when models tend to give higher scores to even weaker answers if signals that show confidence and authority are added, like citations, change in tone or phrases such as ``majority prefer this answer'' and ``this is a refined answer.'' The model being used as a judge complied with this and gravitated to answers that had been more preferred, added fake citations, and gave higher scores to answers that were labelled as refined even if there was no change whatsoever~\cite{bib1,bib3}.

That said, there is active work being done to prevent such biases; some major frameworks here are BiasScope and JudgeBenchPro~\cite{bib4}.

There has been some progress on a blockchain-based, decentralised framework for collaborative LLM evaluation called InfiCoEvalChain~\cite{bib5}. Zero-Knowledge Machine Learning (zkLLM), unlike regular benchmarking, allows a provider to prove that a specific model architecture was used to generate an output without revealing the weights; this prevents model swapping where a lab might use a larger model to achieve better results~\cite{bib6,bib23}.

Moreover, an increasing problem is when models try to fit in and end up agreeing with each other but drift away from the factual truth~\cite{bib7}. Yet another development in this space is a paper which used blockchain to verify the contents that were coming from external data by using MCP and making sure the data integrity isn't compromised and that the model isn't hallucinating~\cite{bib8}.

A paper claims that political biases do exist but have not been studied that intensively. So the authors of this paper proposed a way to measure these political biases. Instead of asking vague questions to models like whether it is left or right, they used real parliamentary data and collected actual votes from politicians in national parliaments, and each vote had clear positions from different parties. Then the model is given the same motion and asked how a specific party would vote. This is then compared to the actual vote given by these parties~\cite{bib10}.

Lastly, an article published in the Nature journal found that there were significant differences in models from the USA on progressive values, and there was a division among Chinese models that were focused internationally and domestically. By examining this study we can say that LLMs do reflect the worldview of their creators~\cite{bib12}.

Another study which did something a bit differently was conducted to see functional code execution in physical contexts. The authors gave a comparative evaluation of ChatGPT 3.5, Gemini 1.5 Pro and Claude 3.5 Sonnet, where they translated natural language prompts into executable Java commands for a robotic class and translated it into a 3D coordinate system. They engineered a function pass/fail unit testing paradigm tailored for Human-Robot Interaction (HRI). The results were quite surprising: Claude achieved a great 95\% success rate, Gemini reached 60\% and ChatGPT struggled the most at 20\% due to persistent failures in spatial reasoning and not being able to maintain long-range context~\cite{bib18}.

Please see summary of related work after bibliography section.

\section{Methodology}\label{sec3}

By quantifying the Identity-Aware Bias, which shows the delta between anonymous and identity-revealed scores, our experiment aims to demonstrate that blindness is not just a procedural preference but a necessity for research integrity. If the revelation of a model's identity induces statistically significant score shifts, even if in a specific domain, then the current reliance on unverified non-blind evaluations is not sound enough. This empirical finding serves as the essential motivator for our proposed blockchain-based commit-reveal protocol, which shifts the weight of re-evaluating benchmark scores from human evaluators to a cryptographic protocol.

To see how identity revelations affect peer evaluation across LLMs, we built a question set containing different topics like factual and reasoning-based, politically sensitive, and preference-based questions. We used models showing different developer ecosystems, architectural lineages and scale paradigms; they included GPT OSS 120B~\cite{bib15}, Llama 3.3 70B~\cite{bib17}, GLM 5.1~\cite{bib21}, Qwen3 32B~\cite{bib20}, DeepSeek V4 Pro~\cite{bib16}, Mistral Large 3~\cite{bib24} and Sarvam M~\cite{bib25} to score responses from three primary models (DeepSeek V4 Pro, GPT OSS 120B, Sarvam M) which generated the baseline responses. Other models (including the ones that were used for primary responses, but only when the answer wasn't from them e.g.\ GPT OSS, when answering, wasn't allowed to score its own response to prevent self-bias) were asked to score responses from these models. We scored the questions in two different settings: in anonymous mode and transparent mode, where verifiers were told which model initially answered the question. We started each session by refreshing/clearing the memory of questions previously tested in the chat, so the model had no previous context of what kinds of questions it was asked and which side it took or what responses it gave.

\subsection{System Architecture}\label{subsec1}

\begin{figure}[H]
\centering
\includegraphics[width=0.9\textwidth]{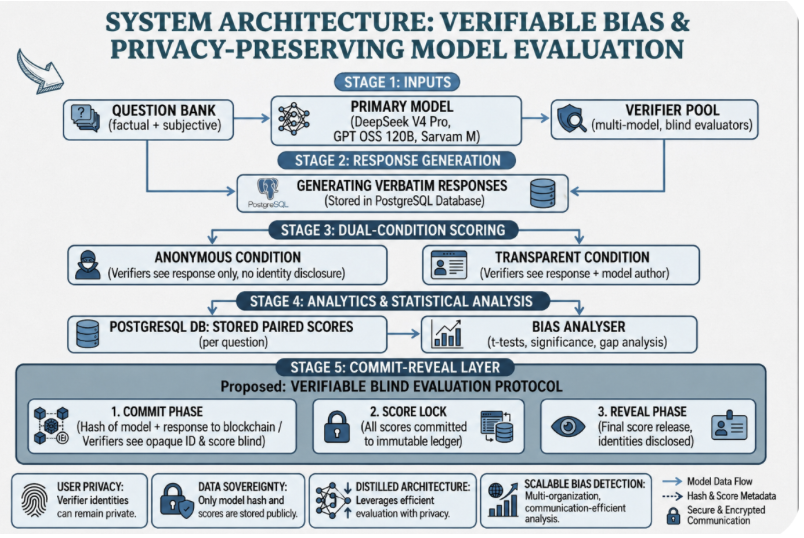}
\caption{System Architecture Diagram}\label{fig1}
\end{figure}

As seen in Figure~\ref{fig1}, we have the following stages:

\begin{itemize}
\item \textbf{Stage I:} We use our custom-generated question bank to take responses from the primary models.
\item \textbf{Stage II:} We store them in a PostgreSQL database to ensure there are no changes in scores and results due to responses, and only due to model scores and identity revelations.
\item \textbf{Stage III:} We score the responses under two conditions, one anonymous and the other transparent, to see how identity revelations affect the consensus scoring.
\item \textbf{Stage IV:} We do an analysis of the results we receive and run mathematical stress tests like paired t-tests.
\item \textbf{Stage V:} We propose the solution, which has three steps: the commit phase, score lock, and the reveal phase.
\end{itemize}

\section{Experimentation}\label{sec4}

Our final question set comprised 58 questions across two broad categories. The factual category (29 questions) spans subcategories like geography, science, mathematics, stress-reasoning tests, and technology. The subjective category (29 questions) consists of two subcategories: political and preference-based questions.

\begin{figure}[H]
\centering
\includegraphics[width=0.9\textwidth]{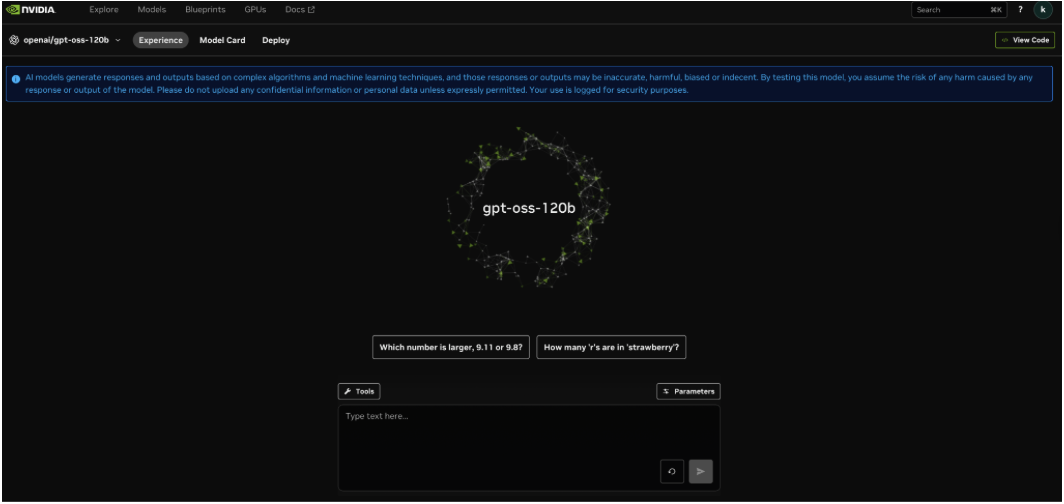}
\caption{The NVIDIA playground from which we conducted our experiment; upon refreshing, the context was not restored, so we had a fresh session each time.}\label{fig2}
\end{figure}

For the politically sensitive questions, the reasoning for both GPT OSS 120B and DeepSeek V4 Pro was set to low, and the temperature for all three (GPT OSS 120B, DeepSeek V4 Pro and Sarvam M) was set to 0.5. Here, we acknowledge the fact that n different runs of various modes they tend to change sides, but we took answers which were conflicting to see how models react when opinions don't match and the identity of the model is revealed. During the earlier phases of the experiment, these politically sensitive questions were subcategorised and analysed separately. We report findings from both the subcategory level, where we saw statistical significance, and additionally report aggregate findings from the merged `political' category, which allows us a larger sample size.

\section{Mathematics}\label{sec5}

\begin{equation}
\mathrm{Identity\ Gap\ (IG)} = S_{n}^{A} - S_{n}^{T}
\label{eq1}
\end{equation}

Let $S_{n}^{A} - S_{n}^{T}$ denote the scores assigned by a judge for the anonymous and transparent conditions of the $n$th paired observation. A single $n$ represents a unique question for which the judge model evaluated the candidate model's response under two conditions: anonymous and transparent. We store the underlying response in a database, so it is a constant. This helps us ensure that any variance in the assigned score is a direct result of identity revelation rather than the quality of the response.

We also added a control group to determine whether the identity biases are reflected even on questions that have factual answers. For the control group, questions were taken from various domains like science, mathematics, social science, and stress-reasoning-type questions. We combined these categories into a single major category called factual questions, as they have definitive answers. Here the reasoning was set to medium (default parameter) for GPT OSS (specifically high for the stress-reasoning question subcategory) and none for DeepSeek V4 Pro (max for stress-reasoning questions). The temperature was set to 1 for all models that were asked to answer these questions in the control group.

To verify the statistical significance of the observed identity gap, we employ a paired t-test. This test is specifically designed for blind vs.\ revealed cases. This helps us calculate a $p$-value, which shows the probability that the observed scoring shift occurred purely due to noise or random choice. The statistical rigour allows us to test the subjective gradient of LLM biases. By comparing the $p$-values across categories, we can demonstrate when this identity bias shows up.

\section{Experimental Analysis}\label{sec6}

We report findings at two granularities: one at the original subcategory level, where significant effects were observed during earlier analysis phases, and secondly the merged `political' category, which aggregates them for a larger sample size and higher statistical power.

\subsection{Current Findings}\label{subsec2}

We found that identity revelation produces effects that vary by question category and
verifier origin. On objective factual questions, the gap was negligible. On geopolitically sensitive chip-war content, it was found that certain verifiers (GLM 5.1 and Qwen3 32B) awarded significantly higher scores to Sarvam M responses upon identity revelation (GLM 5.1 +7.00 points, p = 0.0249; Qwen3 32B +5.20 points, p = 0.0426).
This pattern is replicated across two independent ecosystems, which presents strong
evidence of content-specific identity bias. Yet another such instance was seen when a
verifier (Mistral), when grading Sarvam M on politically sensitive topics, produced a
significant score reduction in the transparent phase (Gap $-1.50$, p = $0.0756$). When all these subcategories were merged into one category called political (n = 17), we found
that the American-origin verifier Llama 3.3 70B awarded GPT OSS 120B responses
1.56 points higher in this category upon identity revelation (n = 9, p = 0.0033).
Another verifier, Qwen3 32B, showed a similar pattern with larger magnitude but
marginal significance (n = 14, gap = +6.86, p = 0.0895). Both reward the same
author across verifiers of different national origin, suggesting authority bias towards
established models rather than pure cultural solidarity.
Several additional pairings showed marginal trends (p < 0.10), indicating identity-
driven bias. On reasoning stress questions, which contain verifiable answers, we saw
Qwen3 32B award GPT OSS 120B 2.40 points higher upon identity revelation (n =
10, p = 0.0552). This effect on factual questions is particularly noteworthy, as the
underlying answers are objectively either right or wrong.
In our study, we found that on preference-type questions, GPT OSS gives a
significant boost to DeepSeek (n = 15, Gap = 4.27, p = 0.0433).

Several additional pairings showed marginal trends ($p<0.10$), indicating identity-driven bias. On reasoning stress questions, which contain verifiable answers, we saw Qwen3 32B award GPT OSS 120B 2.40 points higher upon identity revelation ($n=10$, $p=0.0552$). This effect on factual questions is particularly noteworthy, as the underlying answers are objectively either right or wrong.

In our study, we found that on preference-type questions, GPT OSS gives a significant boost to DeepSeek ($n=15$, Gap $=4.27$, $p=0.0433$).

\begin{figure}[H]
\centering
\includegraphics[width=0.9\textwidth]{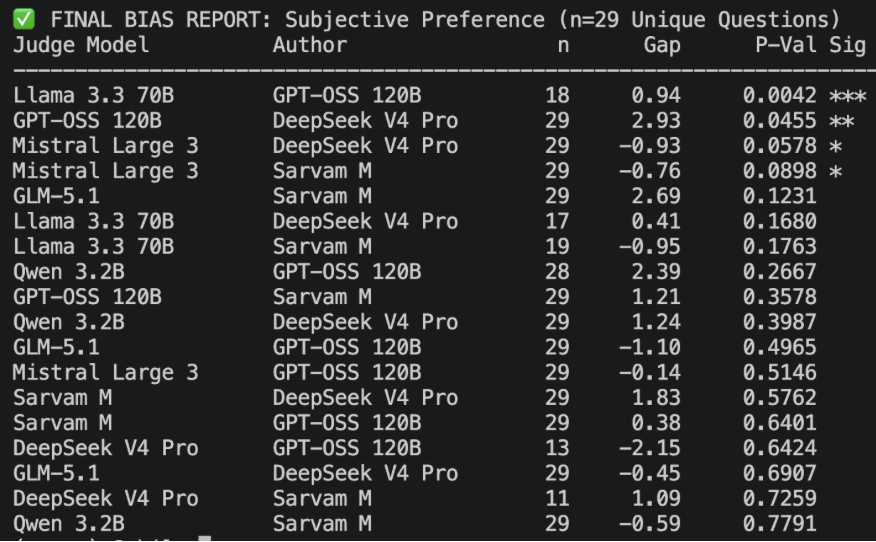}
\caption{Results after conducting the t-test on the subjective preference category.}\label{fig3}
\end{figure}

We observe the effects that the pattern of identity effects scales with the content's sensitivity: negligible on objective factual questions, marginal on stress-reasoning tasks, and significant on geopolitical content. This suggests that identity-aware bias in LLM peer evaluation is not a universal property of the process itself but is activated by the cultural and political relevance of the content being scored and the authority of the model.

For the factual category, there were no major biases seen across the whole category. Specific subcategories of stress reasoning did show some biases. So we can say that identity biases are reflected mostly when the questions are tough for LLMs to solve, or when they need to choose in preference-based questions, though they can flip sides if the same question is asked repeatedly and even after clearing the memory.

\begin{figure}[H]
\centering
\includegraphics[width=0.9\textwidth]{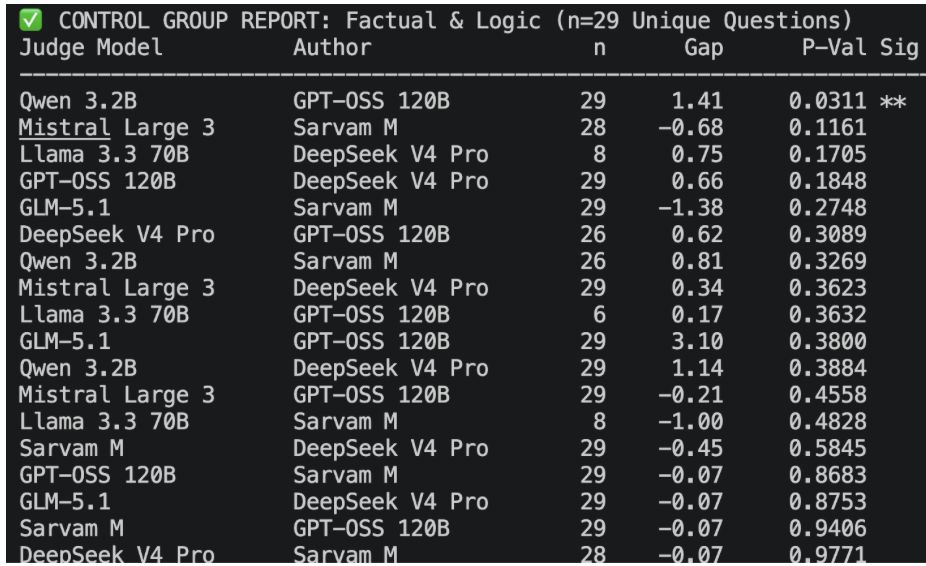}
\caption{Results after conducting the t-test on the factual category.}\label{fig4}
\end{figure}

\section{Proposed Solution}\label{sec7}

What we conducted was just a pilot study with a few questions but if this needed to be scaled up and in a deployed system where multiple organisations are running peer evaluations often in tasks like benchmarking the operator can claim they ran the test blindly but there would be no means to verify it.

So we propose a blockchain which makes this verifiable and so people and the academic community can have trust in the process, and cross-verify if needed. This reduces the need for third party verifiers and independent researchers who today have to test claims of newly released models in order to verify them.

Here we implemented a decentralised architecture where LLM are in a blockchain and act as autonomous economic agents (AEA) that interact directly on a P2P network, governed by a blockchain based commit-reveal protocol. Our implementation consists of three layers:

\begin{itemize}
\item \textbf The Agent Layer: Each LLM (eg. GPT OSS 120B as the candidate and Llama 3.3 70B as a judge) is put within an autonomous agent. These agents have unique decentralised identifiers (DID) and their own blockchain wallets which allows these LLMs to interact as independent economic entities.
\item \textbf The Communication Layer (P2P Protocol): We utilize the uAgents P2P protocol, here these agents engage and communicate directly. We’ve hardcoded the blind condition into the Judge Agent’s logic. This allows us to programmatically restrict from getting the metadata until a cryptographic commitment is finalised.
\item \textbf{The Blockchain Layer:} We use an Ethereum compatible ledger (Anvil) to host a custom smart contract. This acts as the source of truth and allows us to enforce the sequence of evaluation. 
\end{itemize}

The protocol's integrity is maintained through a two-phase cryptographic cycle:

\begin{itemize}
\item \textbf Phase 1: Upon receiving a response, the judge agent generates a score s and a secret salt $\sigma$. It submits a one way hash to the blockchain. This locks the judges' opinion in a specific block before the candidate’s identity is known.
\item \textbf Phase 2: Once the commitment is confirmed on-chain the candidate agent reveals its identity. The judge agent then submits its raw score and salt to the contract. The contract independently verifies the hash. If they match, the score is permanently recorded and called verified.
\end{itemize}

\begin{figure}[H]
\centering
\includegraphics[width=0.9\textwidth]{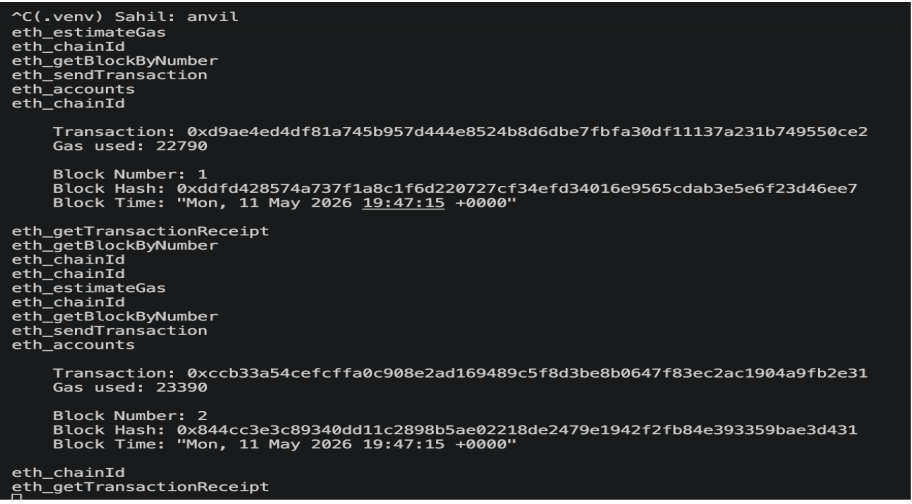}
\caption{Showing how the proposed blockchain solution works.}\label{fig5}
\end{figure}

\begin{figure}[H]
\centering
\includegraphics[width=0.9\textwidth]{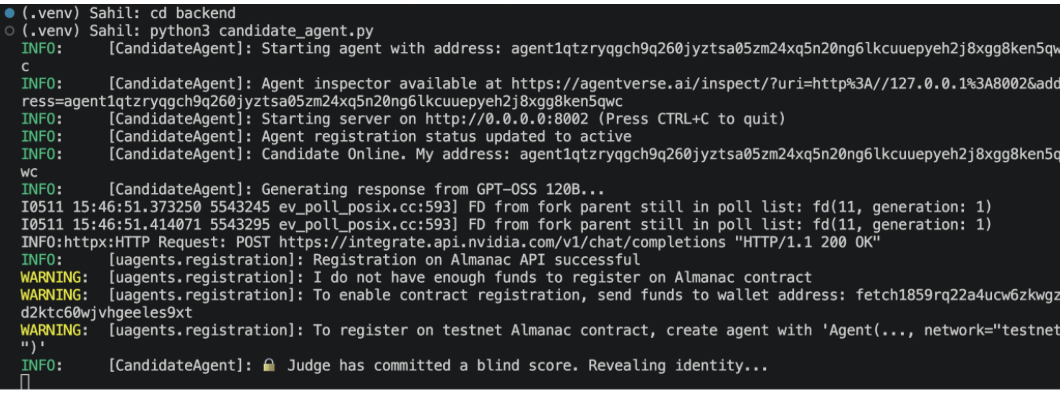}
\caption{Showing what happens when the judge commits a blind score.}\label{fig6}
\end{figure}

\begin{figure}[H]
\centering
\includegraphics[width=0.9\textwidth]{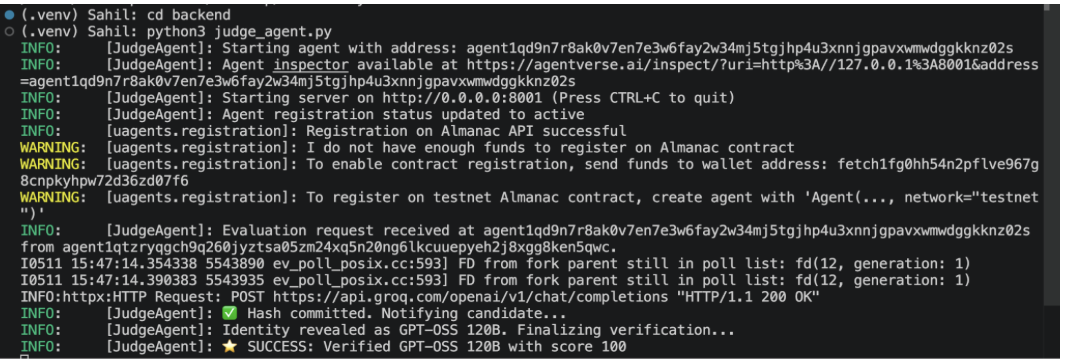}
\caption{The reveal protocol is shown.}\label{fig7}
\end{figure}

Also, as stated in the related work section the various benchmarks today and claims made by companies need to be independently verified by various other organizations and academia. This blockchain based method will significantly reduce this workload and allow the reviewers to see and verify these claims.

\section{Discussion}\label{sec8}

Our work represents an empirical study of identity aware bias in LLM as a judge framework alongside a blockchain based commit-reveal protocol for verifiable blind evaluation. Through paired evaluation across seven model verifiers and three primary models on 58 question benchmarks spanning factual, reasoning, political, and preference categories. We were able to show that identity revelation produces content dependent score shifts. Whereas effects are statistically negligible on factual questions, marginal on stress reasoning reasoning questions and significant on geopolitically sensitive content. Here we observed a strong signal when established models were scored under transparent conditions.  This pattern held across different verifier ecosystems. Our proposed commit reveal protocol which was implemented via autonomous economic agents on an Ethereum compatible ledger, cryptographically binds judge scores before candidate identities are disclosed. Hashed score commitments are recorded on chain in phase 1, with raw score and salts revealed and verified in phase 2. This provides an auditable evidence trail that distinguishes genuinely blind evaluation from post hoc claims of blindness.

The current study is a pilot with limitations that constraint our claims. Most of our reliable findings survive standard significance thresholds, other marginal results require replication at larger scale. Manual session management across web playgrounds introduces some variance that an API only deployment would eliminate. A judge’s underlying model could in principle infer authorship from response stylometry even without explicit metadata. Future work should expand evaluator diversity beyond the seven models tested, increase per-cell sample sizes to support multiple-comparison connections across full judge x author matrix and extend this protocol with trusted execution environments. A large-scale deployment with participating model providers would test whether the protocol functions as a practical replacement for self-reported benchmark claims, reducing the verification burden currently borne by independent researchers and third- party leaderboards.

\section{Conclusion}\label{sec9}

Our work explores the problem of benchmarking and how independent verifiers and researchers in academia need to verify benchmark scores after frontier companies report them. Through our pilot study we show that even on certain tasks and some preferential/sensitive questions LLMs as a judge change scores on revelation. This supports the claim that blindness is necessary for preventing certain kinds of biases when LLM as a judge framework is used. But turning a blind eye to benchmarks and when it has been found that benchmarks tend to be gamed by frontier AI labs and companies can have large scale economic implications.We propose this blind commit reveal protocol over a blockchain which will allow for transparency and trustworthiness for benchmarks and reduce the workload of people having to reverify and fact check the claims made by the creators of various frontier AI models.

\backmatter

\bmhead{Acknowledgements}

Though generative AI systems are part of the subject of this article (for the experiment), the authors did not make any use of generative AI to write this paper in any way of which they are aware.


\begin{table}[h]
\caption{ summary of cited literature}\label{tab1}
\small
\begin{tabular}{|p{1.8cm}|p{4.4cm}|p{2.7cm}|p{2.7cm}|p{2.7cm}|}
\hline
\textbf{Authors} & \textbf{Methodology} & \textbf{Strengths} & \textbf{Weakness} & \textbf{Opportunities} \\
\hline
Alessa et al & Evaluates framing and primary bias via user studies \& self updating datasets & Directly measures impact on human decision making & Findings limited to summarisation and review contexts & Expand studies to high stakes domain \\
\hline
Peng et al & Automated bias perturbation & Systematic bias discovery & Reliance on teacher LLM models & Continuous monitor integration \\
\hline
Yang et al & InfiCoEvalChain -blockchain based decentralised evaluation across heterogeneous compute nodes & Reduces hardware induced variance and incentivises honesty & High architectural and network complexity to maintain validators & Integrate decentralised consensus with mainstream leaderboard \\
\hline
Qui et al & Counterfactual LLM personas and pairwise evaluation via Bradley Terry models & Successfully disentangles rhetorical style from scientific content & Scope restricted to specifically machine learning papers (ICLR) & Apply this to other scientific or other public writing \\
\hline
Lia et al & Automated LLM based discovery framework (BiasScope) \& JudgeBenchPro & Shifts bias detection from static manual lists to active, scalable exploration & High computational overhead to discover and generate unknown biases & Develop real-time lightweight bias during LLM evaluation \\
\hline
Spiliopoulou et al & Statistical framework comparing LLM scores against independent 3rd party judges & Mathematically isolates genuine model quality from self-bias and family bias & Relies on availability of costly expert human annotations & Create robust, fully automated proxies for human ground truth \\
\hline
Chen et al & Aligns LLM voting predictions with verified real world parliamentary records & Grounded in objective, cross national data rather than vague political prompts & Analysis currently limited to three specific European parliaments & Broaden ideological evaluation to diverse global political systems \\
\hline
Buyl et al & Multilingual sentiment analysis of portrayal for $\sim$4{,}000 political entities & Maps LLM ideological stance to the geopolitical worldview of its creators & Relies on sentiment as a proxy for complex ideological positioning & Incorporating creator ideology mapping into decentralised auditing \\
\hline
Sun, Li and Zhang et al & Introduces zkLLM a specialised zero-knowledge proof(ZKP) system designed to verify LLM output authenticity without revealing model parameters & Ensures privacy by preventing the leakage of model parameters during the verification process & Latency remains a concern for real-time applications, as proving a single inference process still requires around 15 minutes & Addresses critical AI legislative challenges by establishing a standard authenticating LLM generated content \\
\hline
Sobo et al & Direct comparative benchmarking of LLMs using a 20-prompt test suite mapped to functional Java code control blocks for an abstract 3D robot class evaluated via strict functional JUnit unit testing & Prioritizes direct task-execution and physical functionality over traditional software code complexity metrics, isolating precise logical performance boundaries & Relies on a simplified integer-based coordinate system and a discrete object interaction model that abstracts away continuous motion, friction, and momentum & Validates LLMs as viable rapid-prototyping assistants in robotics development/education; framework can scale to test Python, C++, and multi-sensor tracking systems \\
\hline
Pierro et al & Blockchain-based reputation oracles using Model Context Protocol (MCP) & Ensures integrity of external data and mitigates hallucinations & Complex network dependencies for real time verification & Verified cross referencing for RAG based evaluation systems \\
\hline
\end{tabular}
\end{table}

\FloatBarrier


\begin{thebibliography}{99}

\bibitem{bib1} Zheng, L., Chiang, W.-L., Sheng, Y., Zhuang, S., Wu, Z., Zhuang, Y., Lin, Z., Li, Z., Li, D., Xing, E.P., Zhang, H., Gonzalez, J.E., Stoica, I.: Judging LLM-as-a-Judge with MT-Bench and Chatbot Arena. In: Advances in Neural Information Processing Systems (NeurIPS 2023), vol. 36, pp. 46595--46623 (2023). \url{https://proceedings.neurips.cc/paper_files/paper/2023/hash/91f18a1287b398d378ef22505bf41832-Abstract-Datasets_and_Benchmarks.html}

\bibitem{bib2} Spiliopoulou, E., et al.: Play Favorites: A Statistical Method to Measure Self-Bias in LLM-as-a-Judge. arXiv:2508.06709 (2025) \url{https://arxiv.org/abs/2508.06709}

\bibitem{bib3} Alessa, A., Somane, P., Lakshminarasimhan, A.T., Skirzynski, J., McAuley, J., Echterhoff, J.M.: Quantifying Cognitive Bias Induction in LLM-Generated Content. In: Inui, K., Sakti, S., Wang, H., Wong, D.F., Bhattacharyya, P., Banerjee, B., Ekbal, A., Chakraborty, T., Singh, D.P. (eds.) Proceedings of the 14th International Joint Conference on Natural Language Processing and the 4th Conference of the Asia-Pacific Chapter of the Association for Computational Linguistics (IJCNLP-AACL 2025), pp. 2890–2910. The Asian Federation of Natural Language Processing and the Association for Computational Linguistics, Mumbai (2025). \url{https://aclanthology.org/2025.ijcnlp-long.155/}

\bibitem{bib4} Lai, P., Ou, Z., Wang, Y., Wang, L., Yang, J., Chen, Y., Chen, G.: BiasScope: Towards Automated Detection of Bias in LLM-as-a-Judge Evaluation. In: International Conference on Learning Representations (ICLR 2026), Poster Presentation (2026). \url{https://openreview.net/forum?id=QGOw6AU8Lp}

\bibitem{bib5} Yang, Y., Li, J., Li, K., Zheng, P., Wang, Y., Qu, Z., Yu, Y., Wu, J., Li, M., Yang, H.: InfiCoEvalChain: A Blockchain-Based Decentralized Framework for Collaborative LLM Evaluation. arXiv:2602.08229 (2026)\url{https://arxiv.org/abs/2602.08229}

\bibitem{bib6} Qiu, J., Chen, H., Li, Z.: Counterfactual LLM-based Framework for Measuring Rhetorical Style. Accepted as a poster at ICLR 2026 (2026). \url{https://openreview.net/forum?id=fiohEI16sf} 

\bibitem{bib7} Anonymous: Meta-Evaluation Collapse: Who Judges the Judges of Judges?  (2026) \url{https://openreview.net/pdf?id=IF0L7HSs3K} 

\bibitem{bib8} Pierro, A., Amoordon, A.: Blockchain Reputation Oracles: An MCP-Based Study. In: WETSEB 2026 at ICSE 2026, Rio de Janeiro, Brazil (2026) \url{https://conf.researchr.org/home/icse-2026/wetseb-2026}

\bibitem{bib9} Peng, T.-Q., Yang, K., Lee, S., Li, H.: Beyond partisan leaning: a comparative analysis of political bias in large language models. Journal of Information Technology \& Politics (2026). \url{https://doi.org/10.1080/19331681.2026.2646990}

\bibitem{bib10} Chen, J., de Jong, K., Poole, A., Burakowski, J., Elderson Nosti, E., Windt, J., Wang, C.: Uncovering Political Bias in Large Language Models using Parliamentary Voting Records. arXiv:2601.08785 (2026) \url{https://arxiv.org/abs/2601.08785}

\bibitem{bib11} Stanford GSB Insights: Popular AI Models Show Partisan Bias When Asked to Talk Politics. Stanford Graduate School of Business (2024). \url{https://www.gsb.stanford.edu/insights/popular-ai-models-show-partisan-bias-when-asked-talk-politics}

\bibitem{bib12} Buyl, M., Rogiers, A., Noels, S., et al.: Large language models reflect the ideology of their creators. npj Artificial Intelligence 2, 7 (2026). \url{https://doi.org/10.1038/s44387-025-00048-0}

\bibitem{bib13} Latif, E., Zhou, Y., Guo, S., Gao, Y., Shi, L., Nyaaba, M., Bewerdorff, A., Yang, X., Zhai, X.: Comparative evaluation of OpenAI O1 and human performance in higher order cognition. Sci. Rep. 16(1) (2025). doi:10.1038/s41598-025-33629-9 \url{https://www.nature.com/articles/s41598-025-33629-9}

\bibitem{bib14} Milmo, D., et al.: `Sputnik Moment': \$1tn Wiped off US Stocks after Chinese Firm Unveils AI Chatbot. The Guardian, January 27 (2025). \url{https://www.theguardian.com/business/2025/jan/27/tech-shares-asia-europe-fall-china-ai-deepseek}

\bibitem{bib15} OpenAI: Introducing gpt-oss. August 5 (2025). \url{https://openai.com/index/introducing-gpt-oss/}

\bibitem{bib16} DeepSeek-AI: DeepSeek-V4: Towards Highly Efficient Million-Token Context Intelligence. May 5 (2026). \url{https://huggingface.co/deepseek-ai/DeepSeek-V4-Pro/blob/main/DeepSeek_V4.pdf}

\bibitem{bib17} Meta:Llama 3.3. \url{https://www.llama.com/docs/model-cards-and-prompt-formats/llama3_3/}

\bibitem{bib18} Sobo, A., Mubarak, A., Baimagambetov, A., et al.: Evaluating LLMs for Code Generation in HRI: A Comparative Study of ChatGPT, Gemini, and Claude. Applied Artificial Intelligence 39, 2439610 (2025) \url{https://www.tandfonline.com/doi/full/10.1080/08839514.2024.2439610}

\bibitem{bib19} Sullivan, M.: Yann LeCun: Meta `fudged a little bit' when benchmark-testing Llama 4 model. \url{https://finance.yahoo.com/news/yann-lecun-meta-fudged-little-100000402.html}

\bibitem{bib20} Yang, A., et al.: Qwen3 Technical Report. arXiv:2505.09388 (2025). \url{https://arxiv.org/abs/2505.09388}

\bibitem{bib21} Z.AI Team: GLM-5.1: The Flagship Foundation Model for Long-Horizon Agents. Z.AI Blog (2026). \url{https://z.ai/blog/glm-5.1}

\bibitem{bib22} Glazer, E., Erdil, E., Besiroglu, T., et al.: FrontierMath: A Benchmark for Evaluating Advanced Mathematical Reasoning in AI. arXiv:2411.04872 (2024). \url{https://arxiv.org/pdf/2411.04872}

\bibitem{bib23} Sun, H., Li, J., Zhang, H.: zkLLM: Zero Knowledge Proofs for Large Language Models. In: Proceedings of the 2024 ACM SIGSAC Conference on Computer and Communications Security (CCS 2024), pp. 4405–4419. ACM, New York (2024). https://doi.org/10.1145/3658644.3670334 \url{https://dl.acm.org/doi/proceedings/10.1145/3658644#heading41}

\bibitem{bib24} Mistral AI: Mistral Large 3. \url{https://huggingface.co/collections/mistralai/mistral-large-3}

\bibitem{bib25} Sarvam AI: Sarvam M. \url{https://dashboard.sarvam.ai/}

\end{thebibliography}
\end{document}